\documentclass[10pt,twocolumn,letterpaper]{article}

\usepackage[pagenumbers]{cvpr} 

\usepackage{amsmath}
\usepackage{amsfonts}
\usepackage[dvipsnames]{xcolor}
\usepackage{csquotes}
\usepackage{tcolorbox}
\tcbuselibrary{breakable}
\tcbuselibrary{skins}
\usepackage{enumitem}
\newtcolorbox{promptbox}[1][]{
    title={#1},
    colframe=gray!15,
    colbacktitle=gray!15,
    colback=gray!5,
    breakable=true,
    fonttitle=\bfseries\large\color{black},
}
\usepackage{booktabs}
\usepackage{multirow}
\usepackage{arydshln}
\usepackage{tabularray}
\usepackage[accsupp]{axessibility}  

\definecolor{cvprblue}{rgb}{0.21,0.49,0.74}
\usepackage[pagebackref,breaklinks,colorlinks,allcolors=cvprblue]{hyperref}

\def\paperID{} 
\def\confName{CVPR}
\def\confYear{2026}

\title{LookasideVLN: Direction-Aware Aerial Vision-and-Language Navigation}

\author{
Yuwei Ning\textsuperscript{1,2}\textsuperscript{\dag}
\
Ganlong Zhao\textsuperscript{3,4}\textsuperscript{\dag}
\
Yipeng Qin\textsuperscript{5}
\
Si Liu\textsuperscript{6}
\
Yang Liu\textsuperscript{1}
\
Liang Lin\textsuperscript{1,2,7}
\
Guanbin Li\textsuperscript{1,2,7$^*$}
\\
\textsuperscript{1}Sun Yat-sen University  \qquad \textsuperscript{2}Peng Cheng Laboratory
\qquad  \textsuperscript{3}The Chinese University of Hong Kong
\\ \textsuperscript{4}Centre for Perceptual and Interactive Intelligence \qquad \textsuperscript{5}Cardiff University
\\ \textsuperscript{6}Beihang University \qquad \textsuperscript{7}Guangdong Key Laboratory of Big Data Analysis and Processing
\\
{\tt\small ningyw@mail2.sysu.edu.cn, glzhao@cpii.hk, qinyipeng1991@gmail.com, liusi@buaa.edu.cn,}\\
\vspace{-4pt}
{\tt\small liuy856@mail.sysu.edu.cn, linliang@ieee.org, liguanbin@mail.sysu.edu.cn}
}

\begin{document}
\maketitle
\def\thefootnote{\dag}\footnotetext{Equal Contribution.}
\def\thefootnote{*}\footnotetext{Corresponding author is Guanbin Li.}

\begin{abstract}
Aerial Vision-and-Language Navigation (Aerial VLN) enables unmanned aerial vehicles (UAVs) to follow natural language instructions and navigate complex urban environments.
While recent advances have achieved progress through large-scale memory graphs and lookahead path planning, they remain limited by shallow instruction understanding and high computational cost. In particular, existing methods rely primarily on landmark descriptions, overlooking directional cues—a key source of spatial context in human navigation.
In this work, we propose LookasideVLN, a new paradigm that exploits directional cues in natural language to achieve both more accurate spatial reasoning and greater computational efficiency. LookasideVLN comprises three core components: (1) an Egocentric Lookaside Graph (ELG) that dynamically encodes instruction-relevant landmarks and their directional relationships, (2) a Spatial Landmark Knowledge Base (SLKB) that provides lightweight memory retrieval from prior navigation experiences, and (3) a Lookaside MLLM Navigation Agent that aligns multimodal information from user instructions, visual observations, and landmark-direction information from ELG for path planning.
Extensive experiments show that LookasideVLN significantly outperforms the state-of-the-art CityNavAgent, even with a single-level lookahead, demonstrating that leveraging directional cues is a powerful yet efficient strategy for Aerial VLN.
\end{abstract}    
\section{Introduction}

\begin{figure}
    \centering
    \includegraphics[width=1.0\linewidth]{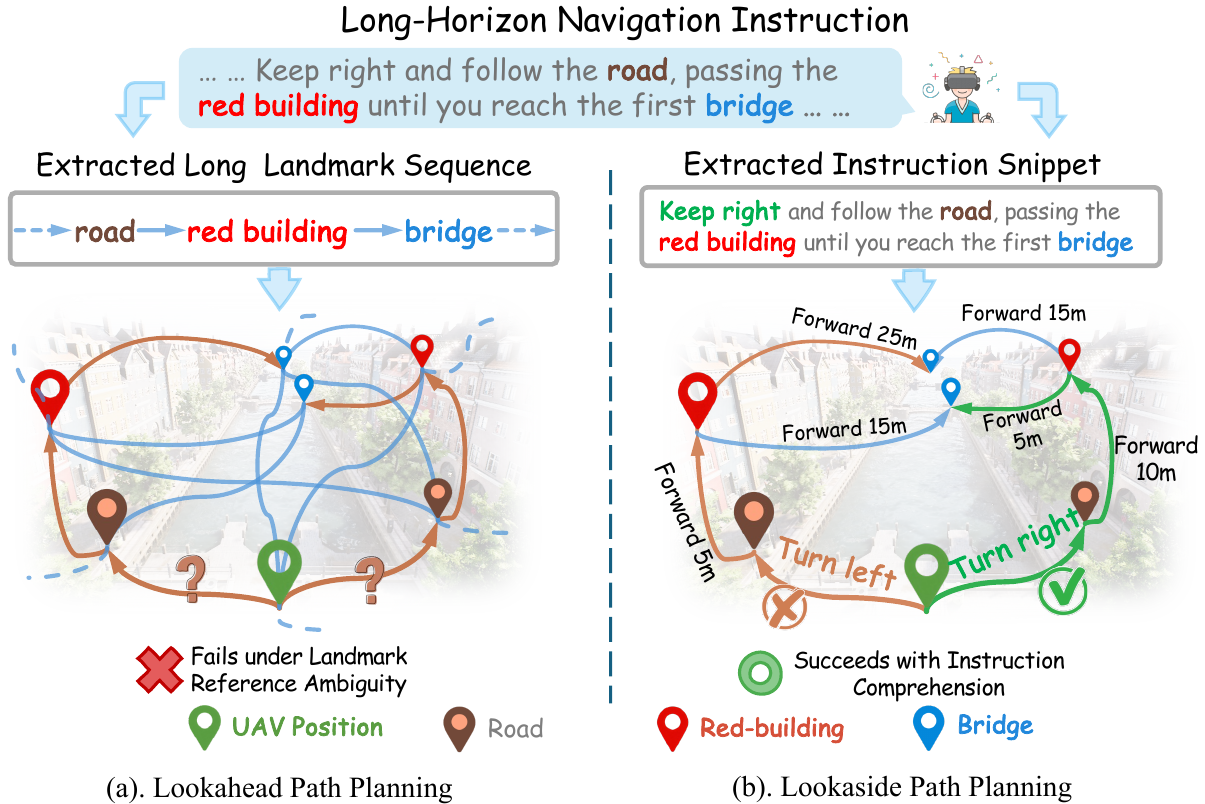}
    \caption{
        Comparison of two path planning paradigms.
        (a) The existing {\it lookahead} paradigm performs path planning by aligning {\it long sequences of landmark} descriptions with UAV observations, which is inefficient and fails to fully disambiguate the landmark descriptions.
        (b) Our \textit{lookaside} paradigm leverages directional cues to enable more efficient and effective path planning.
    }
    \label{fig:insight}
\end{figure}

Aerial Vision-and-Language Navigation~(Aerial VLN)~\cite{liu2023aerialvln, gao2025openfly, wang2024towards, lee2024citynav, fan2022aerial} has emerged as a novel and challenging task that enables unmanned aerial vehicles~(UAVs) to navigate complex aerial environments (\eg, urban scenes) based on natural language instructions, which offers great potential for real-world applications such as urban surveillance, traffic monitoring, disaster response, and aerial delivery.

As a recent addition to the VLN family, Aerial VLN also aims to identify the most appropriate navigation path by aligning UAV's visual observations with natural language instructions. However, large-scale urban environments impose unique challenges absent in small-scale indoor settings.
Specifically, early ground-based VLN methods leverage the {\it uniqueness of landmarks} (a.k.a. objects) to align observations with their textual descriptions~\cite{anderson2018vision, li2023kerm, zhang2024mg, chen2021history, chen2022think}. For instance, a UAV can easily follow ``go to the kitchen'', as most houses contain only one kitchen.
In contrast, this property breaks down in urban environments, where landmark descriptions are often {\it ambiguous} and correspond to multiple instances, e.g., ``tree'', ``wall'' or ``traffic light''. Such ambiguity makes it difficult to infer the correct path by aligning visual inputs with individual landmark descriptions. Consequently, early Aerial VLN methods~\cite{liu2023aerialvln, zhao2025aerial} adapted from ground-based VLN exhibit limited performance.
Recently, \cite{zhang2025citynavagent, shah2023lm} attempt to address this by assuming the {\it uniqueness of landmark sequences} and constructing a global scene graph for sequence-level alignment. While mitigating the issue to some extent, this approach is inefficient due to the heavy computational and memory costs of global graph maintenance in large-scale environments.

In this work, we introduce LookasideVLN, a new {\it lookaside} paradigm that addresses the above limitations by leveraging {\bf directional cues} naturally embedded in user instructions for path planning, in contrast to prior {\it lookahead}-based approaches that depend on long landmark-sequence alignment.
As illustrated in Fig.~\ref{fig:insight}, our key insight is that {\it such directional cues encode rich spatial information, enabling more effective and efficient disambiguation of landmark descriptions in natural language instructions}.
To realize this idea, we design a UAV navigation system that exploits directional cues through a modular architecture comprising three main components:
(1) An \textbf{Egocentric Lookaside Graph (ELG)}.
For the ``egocentric'' design, our rationale stems from the fact that directional cues in natural language instructions (e.g., ``turn left,'' ``fly straight ahead,'' ``go past the building on your right'') are inherently egocentric, grounded in the navigator's viewpoint rather than a global, map-based frame.
Accordingly, we exploit these cues by dynamically constructing an egocentric lookaside graph (ELG) for each instruction. The ELG adopts a hierarchical structure mirroring the sequential landmark descriptions, where each level corresponds to an extracted landmark description and nodes within the same level represent candidate landmarks—visually similar locations retrieved from the UAV's memory.
Owing to this egocentric formulation, the ELG is significantly more efficient than a global scene graph, as it models only instruction-relevant landmarks.
For the ``lookaside'' design, we further model directional relationships along inter-hierarchy edges relative to the UAV’s egocentric orientation, enabling effective spatial reasoning and path planning within a compact graph structure.
(2) A {\bf Spatial Landmark Knowledge Base (SLKB)}, which serves as the UAV's memory for constructing our ELG.
Specifically, it stores descriptions and locations of previously visited landmarks extracted from past navigation experiences.
Our SLKB is both lightweight, as it omits explicit inter-landmark relationships, and efficient, as it enables fast retrieval via a hierarchical (description, location) structure.
Moreover, inspired by {\it Liebig's Law of the Minimum}, we note that the linguistic modality of user instructions inherently constrains the exploitable information for language-vision alignment, rendering detailed visual features used in previous methods~\cite{zhang2025citynavagent, shah2023lm} unnecessary.
(3) A {\bf Lookaside MLLM Navigation Agent}, which employs a multimodal large language model (MLLM) to align the UAV's current observation, user instruction, and the {\it directional landmark information} provided by the ELG for navigation.
This enables comprehensive multimodal alignment, leveraging both semantic and lookaside directional cues in the user instruction to guide navigation more effectively.
Extensive experiments demonstrate the effectiveness of our lookaside paradigm. Specifically, our LookasideVLN, with only a single level of lookahead planning, surpasses the state-of-the-art CityNavAgent~\cite{zhang2025citynavagent}, despite the latter performing full landmark-sequence lookahead over the entire urban scene.

In summary, our contributions include:
\begin{itemize}
    \item We introduce a novel paradigm, namely LookasideVLN, which leverages directional cues in natural language instructions to enhance spatial reasoning and mitigate landmark description ambiguity effectively.
    \item To implement LookasideVLN, we develop a UAV navigation system consisting of three novel components: an Egocentric Lookaside Graph (ELG), a Spatial Landmark Knowledge Base (SLKB), and a Lookaside MLLM Navigation Agent.
    \item Extensive experimental results show that our LookasideVLN achieves state-of-the-art performance, outperforming CityNavAgent~\cite{zhang2025citynavagent} even with a single-level lookahead planning, highlighting the surprising effectiveness of lookaside directional cues.
\end{itemize}

\section{Related Work}

\textbf{Vision-and-Language Navigation.}
Vision-and-Language Navigation~(VLN) is a core task in embodied AI, first introduced by the Room-to-Room (R2R) benchmark \cite{anderson2018vision}, where agents navigate discrete indoor environments. R2R-CE \cite{krantz2020beyond} extends VLN to continuous settings, bridging the gap between simulation and real-world navigation.
Early approaches \cite{fried2018speaker, chen2021history, hong2021vln, guhur2021airbert, chen2022think, kamath2023new, li2023kerm, liu2024volumetric} rely on end-to-end trained models that align language instructions with visual observations to predict actions. 
However, these approaches often suffer from error accumulation during sequential action prediction, causing the agent's trajectory to gradually drift away from the intended route.

Liu et al. \cite{liu2023aerialvln} introduce Aerial VLN, a more challenging benchmark that requires agents to navigate in continuous aerial environments. They adapt the cross-modal attention mechanism from ground-based VLN and employ the look-ahead training paradigm \cite{raychaudhuri2021language}, in place of the previously used student-forcing strategy \cite{anderson2018vision}.
Subsequent efforts have been made toward Aerial VLN. For instance, Zhao et al. \cite{zhao2025aerial} introduce a grid-based view selection mechanism and a top-down map representation to facilitate effective aerial navigation.
Nonetheless, these methods inherit the limitation of error accumulation from ground-based VLN, limiting their effectiveness in long-horizon navigation scenarios.
Other works explore zero-shot LLM-based approaches for Aerial VLN by decomposing long-horizon navigation tasks into sequential subgoals, and developing specialized scene modeling techniques. STMR \cite{gao2024stmr} introduces an LLM-readable semantic top-down map by projecting semantically segmented point clouds into a textual matrix, serving as the agent’s scene perception. However, this top-down representation compresses vertical (height) information, which is crucial for effective aerial navigation. CityNavAgent \cite{zhang2025citynavagent}, inspired by LM-Nav \cite{shah2023lm}, adopts lookahead path planning by incorporating a graph search algorithm over a city-scale scene graph, allowing for more holistic reasoning. Nevertheless, it primarily relies on the semantic similarity between landmark descriptions and visual observations, while ignoring directional cues and lacking a thorough understanding of the instructions. In this work, we address this with a novel {\it lookaside} paradigm.

\vspace{1mm} \noindent
\textbf{LLM for Embodied Planning.}
Recently, Large Language Models~(LLMs) \cite{touvron2023llama, achiam2023gpt, guo2025deepseek, qwen2025qwen25technicalreport, zhao2026navgemini, song2025towards} have demonstrated remarkable capabilities in language understanding and reasoning, attracting widespread interest in their application to robotics. In particular, LLMs have been employed for embodied task planning and motion planning. Some studies leverage the zero-shot or few-shot prediction capabilities of LLMs to generate static plans for embodied agents \cite{kannan2024smart, cao2023robot, di2023towards}, while others enable agents to adapt their plans or actions in response to environmental feedback \cite{chen2024scalable, zhao2023chat}.

More recent work has explored memory-augmented approaches for embodied planning. EmbodiedRAG \cite{booker2024embodiedrag} is the first to integrate Retrieval-Augmented Generation~(RAG) \cite{lewis2020retrieval} into embodied planning, leveraging subgraphs retrieved from scene graphs to reduce run-time computational overhead. RAG-Driver \cite{yuan2024rag} enhances explainable autonomous driving by retrieving driving experiences similar to the current scenario from a memory database.
In the context of indoor VLN, KERM \cite{li2023kerm} and Bao et al. \cite{bao2025enhancing} adopt RAG by retrieving textual descriptions relevant to the agent’s observations from an external image caption dataset \cite{krishna2017visual}. However, relying on external datasets introduces domain gaps, which can hinder robust navigation performance.
In this work, we construct a Spatial Landmark Knowledge Base comprising landmark descriptions and positions, serving as a spatial memory module for the agent.
\section{Method}

\begin{figure*}[t]
    \centering
    \includegraphics[width=1.0\textwidth]{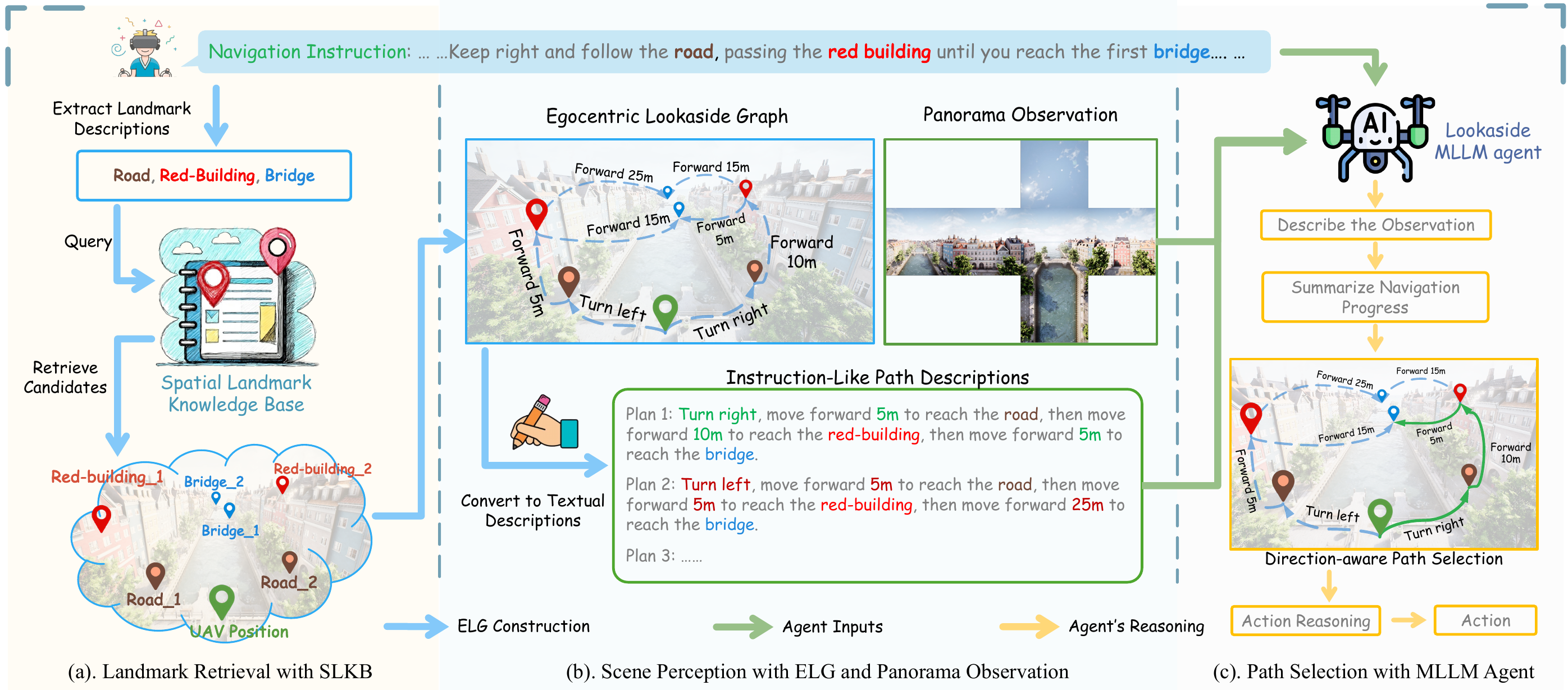}
    \caption{
        Overall framework of LookasideVLN. The agent queries the Spatial Landmark Knowledge Base to construct an Egocentric Lookaside Graph, which is then converted into instruction-like, direction-aware path descriptions. A Lookaside MLLM agent selects the final path based on the alignment among the directional path, instruction snippet, and current observation.
    }
    \label{fig:main_method}
\end{figure*}

In this section, we introduce \textbf{LookasideVLN}, a systematic navigation framework which leverages {\it directional cues} naturally embedded in natural language instructions for path planning, implementing a novel {\it lookaside} paradigm. 
As illustrated in Fig.~\ref{fig:main_method}, LookasideVLN first queries the Spatial Landmark Knowledge Base (SLKB) using landmarks extracted from navigation instructions to retrieve locations for each candidate landmark (Sec.~\ref{sec:lkb}). It then constructs the Egocentric Lookaside Graph (ELG) to model both landmarks and their directional relationships (Sec.~\ref{sec:elsg-construction}). Next, directional paths on the ELG are converted into instruction-like descriptions, enabling the agent to evaluate their alignment with user instructions (Sec.~\ref{sec:path-descriptions}). Finally, to support effective path planning, we introduce a Lookaside MLLM Navigation Agent, which makes path decisions in a robust and explainable manner by jointly reasoning over the scene understanding and navigation instructions (Sec.~\ref{sec:ign-cot}).

\subsection{Spatial Landmark Knowledge Base}
\label{sec:lkb}

Our \textbf{Spatial Landmark Knowledge Base (SLKB)} is an efficient, lightweight, and scalable memory module built from both scene observations and historical navigation records, enables the agent to recall previously observed landmarks and progressively accumulate scene knowledge over time.

\subsubsection{Lightweight and Scalable Design}
\begin{figure}[t]
    \centering
    \includegraphics[width=1.0\linewidth]{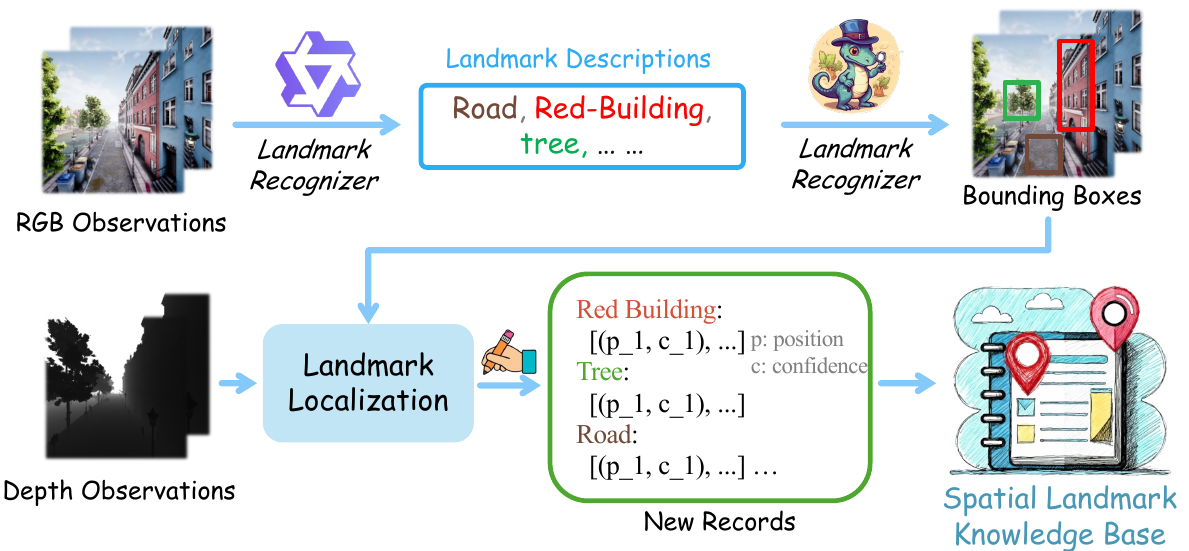}
    \caption{Spatial Landmark Knowledge Base Construction.}
    \label{fig:database_construction}
\end{figure}

To reduce redundancy and achieve efficient retrieval, we design our SLKB $\mathcal{K}$ hierarchically as:
\begin{equation} \label{eq:database}
\mathcal{K} = \big\{ l^{kb}_i : \{p^{kb}_{i,0}, p^{kb}_{i,1}, \dots, p^{kb}_{i,j}\} \big\},
\end{equation}
where $l^{kb}_i$ is the $i$-th landmark description, $p^{kb}_{i,j}$ denotes the $j$-th candidate location for $l^{kb}_i$.
Since our SLKB is a dynamically updated memory, we show below how a new entry is inserted or merged (Fig.~\ref{fig:database_construction}). Historical navigation records are incorporated in the same way.

\vspace{1mm}
\noindent \textbf{Extraction of Newly-observed $l^{kb}$ and $p^{kb}$.} Given an RGB observation $I_{rgb}$ from the scene, we use a MLLM-driven \textit{Landmark Recognizer} $\mathrm{LR}(\cdot)$ to generate descriptions $l^{kb}_i$ for each landmark in $I_{rgb}$.
Next, we use a \textit{Landmark Detector} $\mathrm{LD}(\cdot)$, empowered by GroundingDINO \cite{liu2024grounding}, to predict bounding boxes $\mathcal{B} = \{b_1, b_2, \dots\}$ for $l^{kb}_i$, \ie,
\begin{equation} \label{eq:bbox}
\mathcal{B} = \mathrm{LD}(l^{kb}_i).
\end{equation}
We then filter redundant, overlapping bounding boxes using Non-Maximum Suppression (NMS) to obtain the final bounding box $b$ and compute its 3D location from the corresponding depth observation $I_{depth}$ of $I_{rgb}$. 
Given the center pixel coordinate $(u, v)$ of bounding box $b$ and the mean depth value $\bar{d}$ within the box (computed after removing outliers beyond $2\sigma$ from the depth distribution), the landmark location $p^{kb}_i \in \mathbb{R}^3$ in world coordinates is calculated as:
\begin{equation} \label{eq:landmark_localization} \begin{aligned}
p^{kb}_i = \frac{\bar{d}_i}{\|K^{-1} p^{pixel}_i\|_2} &\cdot R K^{-1} p^{pixel}_i + T,\\
p^{pixel}_i = &[u, v, 1]^\top 
\end{aligned} \end{equation}
where $R \in SO(3)$ and $T \in \mathbb{R}^3$ are the camera's rotation matrix and translation vector, while $K \in \mathbb{R}^{3\times3}$ is the camera intrinsic matrix.
Note that we omit landmark descriptions from user instructions, as they are already covered by those extracted from RGB observations, which also provide the corresponding landmark positions.
Further implementation details are provided in the supplementary materials.

\vspace{1mm}
\noindent \textbf{Insertion or Merging of ($l^{kb}$, $p^{kb}$).} Following the standard database operations, given a new landmark–position pair ($l^{kb}$, $p^{kb}$), our SLKB $\mathcal{K}$ determines whether $l^{kb}$ corresponds to an existing entry. If a similar landmark already exists, the new $p^{kb}$ is merged to update the existing entry of $l^{kb}$. Otherwise, a new record is inserted into the SLKB. This strategy ensures that the knowledge base remains compact and up to date while avoiding redundant entries and preserving spatial consistency across navigation experiences.

\vspace{1mm}
\noindent \textbf{Remark on Compact Landmark Description $l^{kb}$.}
Note that, inspired by {\it Liebig's Law of the Minimum} (a.k.a., the barrel theory), we argue that the linguistic modality of user instructions inherently limits the exploitable information for language–vision alignment, making the detailed visual features used in prior methods~\cite{zhang2025citynavagent, shah2023lm} unnecessary.
Accordingly, we represent each landmark $l^{kb}_i$ textually, reducing memory and computational overhead.

\textbf{}
\subsubsection{Efficient Data Retrieval} \label{sec:lkb-retrieval}
As shown in Fig.~\ref{fig:main_method}, given our hierarchical SLKB $\mathcal{K}$, for each landmark description extracted from the user instruction $\{l^{instr}_i\}_{i=1}^{N}$, we compute its word embedding \cite{li2023towards} and compare it with the embeddings of all landmark descriptions $l^{kb}_j$ in $\mathcal{K}$ using cosine similarity $\textrm{sim}(\cdot)$, retrieving the one with maximum similarity:
\begin{equation}
    l^{ret}_i = \operatorname{argmax}_{l^{kb}_j \in \mathcal{K}} \textrm{sim}\big(\operatorname{emb}(l^{instr}_i), \operatorname{emb}(l^{kb}_j) \big),
\end{equation}
where $\operatorname{emb}(\cdot)$ is the word embedding model. Then, we can efficiently retrieve all positions of each landmark satisfying the description $l^{ret}_i$ as:
\begin{equation}
    \mathcal{L}^{ret} = \big\{ l^{ret}_i: \{ p^{ret}_{i,0}, p^{ret}_{i,1}, \dots \} \big\}_{i=1}^{N}.
    \label{eq:retrived_set}
\end{equation}

\subsection{Egocentric Lookaside Graph}

Navigation instructions typically adopt an egocentric perspective --- such as \textit{``turn left and go to the intersection, then turn right and go to the bridge''} --- where directional cues are essential for disambiguating between multiple candidate landmarks of the same category~(e.g., multiple bridges in the environment). 
In such instructions, directional cues like \textit{``turn right''} are defined relative to \textbf{the agent's orientation upon reaching the preceding key landmark} (i.e. \textit{the intersection} in this case). We refer to these as \textbf{egocentric lookaside directional relationships}, since they depend on the agent’s future orientation after passing key landmarks in the instruction.
To effectively leverage directional cues in navigation instructions, we propose the \textbf{Egocentric Lookaside Graph~(ELG)}, where nodes represent candidate locations of unvisited landmarks mentioned in the instruction, and edges contain egocentric lookaside directional relationships between them.
Unlike previous works \cite{zhang2025citynavagent, shah2023lm} that rely solely on proximity in large-scale scene graphs, our ELG dynamically captures instruction-relevant landmarks and directions to construct small egocentric graphs, where paths can be seamlessly translated into instruction-like descriptions, facilitating semantic scene understanding and effective path planning.

\subsubsection{Egocentric Lookaside Graph (ELG) Construction}
\label{sec:elsg-construction}

Starting from the UAV's current position, we construct the ELG $\mathcal{G}(\mathcal{N}, \mathcal{E})$ by selecting the next $N_{ahead}$ {\it unvisited} landmarks from $\mathcal{L}^{ret}$ (Eq.~\ref{eq:retrived_set}) to form:
\begin{equation} \label{eq:unvis_landmarks}
    \mathcal{L}^{unvis} = \big\{ l^{unvis}_i: \{ p^{unvis}_{i,0}, p^{unvis}_{i,1}, \dots \} \big\}_{i=1}^{N_{ahead}},
\end{equation}
where $l^{unvis}_i$ is the $i$-th unvisited landmark and $p^{unvis}_{i,j}$ is the $j$-th candidate location of $l^{unvis}_i$.
We then construct the ELG hierarchically, where each layer corresponds to a landmark description $l^{unvis}_i$ in sequence, and each node within a layer represents a unique location $p^{unvis}_{i,j}$. The collection of all nodes forms $\mathcal{N}$.
For each pair of successive unvisited landmark layers ($l^{unvis}_{i-1}$ and $l^{unvis}_i$) in the sequence, we create edges between every candidate location $p^{unvis}_{i-1,j}$ of the preceding landmark and every candidate location $p^{unvis}_{i,k}$ of the subsequent landmark.
These edges are then assigned their corresponding egocentric lookaside directional relationships as follows.

\vspace{1mm}
\noindent \textbf{Egocentric Direction Estimation from Location.}
Considering the candidate locations of three consecutive landmarks $(p^{unvis}_{i-1,j}, p^{unvis}_{i,k}, p^{unvis}_{i+1,m})$, we first estimate the agent’s orientation by computing the direction from $p^{unvis}_{i-1,j}$ to $p^{unvis}_{i,k}$. This estimated orientation is then used to determine the egocentric lookaside directional relationship between $p^{unvis}_{i,k}$ and $p^{unvis}_{i+1,m}$.
Specifically, the agent’s orientation upon reaching $p^{unvis}_{i,k}$ is estimated as the unit vector:
\begin{equation}
    \mathbf{p}^{i,k}_{i-1,j} = \frac{p^{unvis}_{i,k} - p^{unvis}_{i-1,j}}{\| p^{unvis}_{i,k} - p^{unvis}_{i-1,j} \|_2 },
\end{equation}
where $\mathbf{p}^{i,k}_{i-1,j} \in \mathbb{R}^3$ represents the agent's viewing direction after transitioning from $p^{unvis}_{i-1,j}$ to $p^{unvis}_{i,k}$.
Then we compute the agent’s horizontal deflection angle $\theta^{i+1,m}_{i,k}$, elevation gain $e^{i+1,m}_{i,k}$, and relative horizontal distance $d^{i+1,m}_{i,k}$ from candidate location $p^{unvis}_{i, k}$ to $p^{unvis}_{i+1, m}$ as follows:
\begin{equation} \label{eq:relative_pos} \begin{aligned}
    \theta^{i+1,m}_{i,k} = \mathrm{hangle}(\mathbf{p}^{i,k}_{i-1,j}, \mathbf{p}^{i+1,m}_{i,k}) \\
    e^{i+1,m}_{i,k} = |(p^{unvis}_{i+1,m} - p^{unvis}_{i,k})_z| \\
    d^{i+1,m}_{i,k} = \|(p^{unvis}_{i+1,m} - p^{unvis}_{i,k})_{xy}\|_2,
\end{aligned} \end{equation}
where $\mathrm{hangle}(\mathbf{v}_1, \mathbf{v}_2)$ \footnote{$\mathrm{hangle}(\mathbf{v}_1, \mathbf{v}_2) = \textrm{atan2}((\mathbf{v}_1 \times \mathbf{v}_2)_z, \mathbf{v}_1 \cdot \mathbf{v}_2 - v_{1x}v_{2x})$} computes the horizontal deflection angle (in the $xy$-plane) from $\mathbf{v}_1$ to $\mathbf{v}_2$, $(\cdot)_z$ extracts the vertical component and $(\cdot)_{xy}$ denotes the horizontal projection of a vector.
We utilize the variables defined in Eq.~\ref{eq:relative_pos} to represent the egocentric lookaside directional relationships between candidate landmarks, enabling the ELG to be seamlessly translated into instruction-like path descriptions for downstream MLLM-based semantic-level path planning.

\subsubsection{Generating Path Descriptions from ELG}
\label{sec:path-descriptions}

To enable the agent to better align instruction understanding with scene comprehension, we convert each possible path in the ELG into an instruction-like description. These descriptions, combined with the navigation instruction, allow the agent to reason and perform path planning in a more intuitive and language-aligned manner.
Given $\theta^{i+1,m}_{i,k}$, $e^{i+1,m}_{i,k}$ and $d^{i+1,m}_{i,k}$ defined in Eq.~\ref{eq:relative_pos}, we describe the motion from $p^{unvis}_{i, k}$ to $p^{unvis}_{i+1, m}$ of the unvisited landmark $l^{unvis}_{i+1}$ as:
\begin{displayquote}
    ``Turn left/right $|\theta^{i+1,m}_{i,k}|$ degrees, move forward $d^{i+1,m}_{i,k}$ meters and ascend/descend $e^{i+1,m}_{i,k}$ meters to reach $l^{unvis}_{i+1}$.''
\end{displayquote}
This detailed form is used for the first unvisited landmark. For later steps, we use a coarser description:
\begin{displayquote}
    ``Turn left/right $|\theta^{i+1,m}_{i,k}|$ degrees and move toward the $l^{unvis}_{i+1}$.''
\end{displayquote}
By traversing all possible paths in the ELG, we generate a set of full path descriptions $\mathcal{P}$, where each entry describes a candidate path. An example from $\mathcal{P}$ is as follows:
\begin{displayquote}
    ``Turn left 30 degrees, move forward 10 meters and descend 4 meters to reach the intersection, then turn right 45 degrees and proceed toward the bridge, ...''
\end{displayquote}
All such descriptions are compiled into a prompt to support semantic-level path planning. We also apply a distance-based pruning strategy to improve reasoning efficiency. The details are provided in the supplementary materials.

\subsection{Lookaside MLLM Navigation Agent}
\label{sec:ign-cot}

We introduce a Lookaside MLLM Navigation Agent that achieves robust and interpretable path planning. As illustrated in Fig.~\ref{fig:main_method}, our MLLM agent processes the navigation instruction $\mathcal{I}$, the path descriptions $\mathcal{P}$ derived from the ELG, and the current panoramic observation $O_t = \{o_{t,i}\}_{i=1}^{6}$ --- comprising front, left, right, back, top, and bottom view images --- to sequentially: 1) generate an Observation Description, 2) summarize the Navigation Progress, 3) perform Direction-aware Path Planning, and 4) conduct Action Reasoning to determine the next action $a_t$.

\subsubsection{Observation Description and Navigation Progress}
At the start of the chain-of-thought reasoning, we prompt the agent to generate an Observation Description, which helps it understand its surroundings by producing a textual summary of the input observation images. In addition, following the approach of prior works \cite{gao2024stmr, zhou2024navgpt}, we also require the agent to summarize its navigation progress, enabling it to assess and track the current stage of the task.

\subsubsection{Direction-aware Path Planning}
To align the agent’s navigation decisions with natural language instructions, we propose a Direction-aware Path Planning strategy. The objective is to select an appropriate direction-aware path description from the candidate set $\mathcal{P}$ on the ELG, guided by the navigation instruction $\mathcal{I}$. Specifically, the agent is instructed to: 1) identify the unvisited landmarks within the candidate paths $\mathcal{P}$; 2) extract the corresponding snippets from $\mathcal{I}$ that describe how to reach these landmarks; and 3) select the most suitable path from $\mathcal{P}$ based on the grounded instruction segments.

\subsubsection{Action Reasoning}
After selecting a path from the ELG, the agent performs an Action Reasoning process to determine the next action $a_t$. This process is guided by the next unvisited landmark on the selected path and further refined using the instruction $\mathcal{I}$ and current visual observation $O_t$. 
Detailed prompts are provided in the supplementary materials.
\section{Experiment}

\begin{table*}[!t]
  \centering
  \setlength{\tabcolsep}{1.2mm}
  \begin{tabular}{lcccccccccc}
    \toprule
    \multicolumn{2}{c}{\multirow{2}{*}{Category}} & \multirow{2}{*}{Method} & \multicolumn{4}{c}{Validataion Seen} & \multicolumn{4}{c}{Validation Unseen} \\
    \cmidrule(lr){4-7} \cmidrule(lr){8-11}
    ~ & ~ & ~ & SR$\uparrow$ & OSR$\uparrow$ & SDTW$\uparrow$ & NE$\downarrow$ & SR$\uparrow$  & OSR$\uparrow$ & SDTW$\uparrow$ & NE$\downarrow$ \\

    \midrule
    \multicolumn{2}{c}{\multirow{2}{*}{Statistical}}
    & Random & 0.0 & 0.0 & 0.0 & 300.8 & 0.0 & 0.0 & 0.0 & \underline{351.0}  \\
    & & Action Sampling & 0.1 & 2.1 & 0.1 & 383.1 & 0.2 & 2.1 & 0.1 & 434.9 \\ 

    \midrule
    \multicolumn{2}{c}{\multirow{3}{*}{Learning-Based}}
    & Seq2Seq & 2.9 & 10.2 & 1.0 & 480.4 &1.1 & 5.6 & 0.3 & 551.8 \\
    & & CMA & 2.3 & 6.5 & 0.8 & 293.5 & 1.6 & 4.4 & 0.5 & 360.7 \\
    & & Zhao'25 & \textbf{7.5} & \underline{12.6} & \textbf{3.3} & \textbf{271.1} & \underline{3.2} & \underline{8.1} & \textbf{1.3} & \textbf{333.5} \\

    \midrule
    \multicolumn{2}{c}{\multirow{1}{*}{Zero-Shot}}
    & LookasideVLN (Ours) & \underline{5.7} & \textbf{26.1} & \underline{1.2} & \underline{278.6} & \textbf{6.4} & \textbf{21.3} & \underline{1.2} & 487.0 \\

    \bottomrule
    \end{tabular}
    \caption{
        Overall performance comparison on the AerialVLN benchmark.
        Bold and \underline{underline} indicate the first and second best results.
    }
    \label{tab:overall-performance-l}
\end{table*}

\begin{table*}[t]
  \centering
  \setlength{\tabcolsep}{1.2mm}
  \begin{tabular}{lccccccccccc}
    \toprule
    \multicolumn{2}{c}{\multirow{2}{*}{Category}} & \multirow{2}{*}{Method} & \multirow{2}{*}{LLM} & \multicolumn{4}{c}{Validataion Seen} & \multicolumn{4}{c}{Validation Unseen} \\
    \cmidrule(lr){5-8} \cmidrule(lr){9-12}
    ~ & ~ & ~ & ~ & SR$\uparrow$ & OSR$\uparrow$ & SDTW$\uparrow$ & NE$\downarrow$ & SR$\uparrow$  & OSR$\uparrow$ & SDTW$\uparrow$ & NE$\downarrow$ \\

    \midrule
    \multicolumn{2}{c}{\multirow{2}{*}{Statistical}}
    & Random & - & 0.0 & 0.0 & 0.0 & 109.6 & 0.0 & 0.0 & 0.0 & 149.7  \\
    & & Action Sampling & - & 0.9 & 5.7 & 0.3 & 213.8 & 0.2 & 1.1 & 0.1 & 237.6 \\ 
    
    \midrule
    \multicolumn{2}{c}{\multirow{3}{*}{Learning-Based}}
    & Seq2Seq & - & 4.8 & 19.8& 1.6 & 146.0 & 2.3 & 11.7 & 0.7 & 218.9 \\
    & & CMA & - & 3.0 & 23.2 & 0.6 & 121.0 & 3.2 & 16.0 & 1.1 & 172.1 \\
    & & LAG & - & 7.2 & 15.7 & 2.4 & 90.2 & 5.1 & 10.5 & 1.4 & 127.9 \\
    
    \midrule
    \multirow{5}{*}{Zero-Shot}
    & \multirow{2}{*}{Indoor} & NavGPT & GPT-4 & 0.0 & 0.0 & 0.0 & 163.5 & 0.0 & 0.0 & 0.0 & \underline{82.1} \\
    & & MapGPT & GPT-4V & 2.1 & 4.7 & 0.8 & 124.9 & 0.0 & 0.0 & 0.0 & 107.0 \\
    \cmidrule{2-12}
    & \multirow{3}{*}{Aerial} & STMR & GPT-4o & 12.6 & \textbf{31.6} & - & 96.3 & 10.8 & 23.0  & - & 119.5 \\
    & & CityNavAgent & GPT-4V & \underline{13.9}  & 30.2 & \underline{5.1} & \underline{80.8} & \underline{11.7} & \underline{35.2} & \textbf{5.0} & \textbf{60.2} \\
    & & LookasideVLN (Ours) & Qwen2.5-VL-72B & \textbf{14.7} & \underline{31.2} & \textbf{5.4} & \textbf{77.1} & \textbf{12.6} & \textbf{36.0} & \underline{3.6} & 100.9 \\

    \bottomrule
  \end{tabular}
  \caption{
        Overall performance comparison on the AerialVLN-S benchmark. Bold and \underline{underline} indicate the first and second best results.
  }
  \label{tab:overall-performance}
\end{table*}

\noindent\textbf{Dataset}
We evaluate our LookasideVLN on the challenging AerialVLN benchmark \cite{liu2023aerialvln}, which comprises 8,446 flight trajectories collected from experienced UAV pilots across 25 diverse city-scale environments in Unreal Engine 4.
AerialVLN emphasizes long-horizon navigation, with an average path length of 661.8 meters.
Each episode consists of an aerial trajectory with a corresponding natural language navigation instruction. The validation set is partitioned into ``Seen'' and ``Unseen'' splits according to the presence of each scene in the training set. AerialVLN-S is a small-scale variant of AerialVLN, consisting of 17 compact scenes.

\noindent\textbf{Metrics}
Following the evaluation metrics of AerialVLN, we assess LookasideVLN’s performance using four standard metrics: \textbf{Success Rate (SR)}, which counts a navigation as successful if the agent stops within 20 meters of the destination; \textbf{Oracle Success Rate (OSR)}, which considers a navigation as oracle success if any point on the trajectory is within 20 meters of the destination; \textbf{Navigation Error (NE)}, defined as the distance between the agent’s stopping point and the destination; and \textbf{Success Rate weighted by Normalized Dynamic Time Warping (SDTW)}, which accounts for both task success and the similarity between predicted and ground-truth trajectories.

\noindent\textbf{Baselines}
We compare LookasideVLN with a range of baselines across four categories.
\textit{Statistical methods}, including \textbf{Random} and \textbf{Action Sampling}~\cite{liu2023aerialvln}, serve to illustrate the size of the solution space and the similarity in action distributions between the training and evaluation splits.
\textit{Learning-based methods} cover traditional end-to-end action prediction models such as \textbf{Seq2Seq}~\cite{anderson2018vision}, \textbf{CMA}~\cite{krantz2020beyond},  \textbf{LAG}~\cite{liu2023aerialvln}, and the method of \textbf{Zhao et al.}~\cite{zhao2025aerial} (denoted as zhao’25 in the tables).
To assess generalization from indoor domains, we also include \textit{zero-shot indoor VLN methods} such as \textbf{NavGPT}~\cite{zhou2024navgpt} and \textbf{MapGPT}~\cite{chen2024mapgpt}. 
Finally, we compare with \textit{zero-shot Aerial VLN methods}, including the state-of-the-art \textbf{STMR}~\cite{gao2024stmr} and \textbf{CityNavAgent}~\cite{zhang2025citynavagent}.
Note that, due to the challenging nature of the AerialVLN dataset, existing zero-shot baselines cannot be directly applied. Thus, we primarily compare against learning-based methods on AerialVLN and include zero-shot baselines in the evaluation on AerialVLN-S.

\noindent\textbf{Implementation Details}
We implement LookasideVLN based on Qwen2.5-VL-72B \cite{Qwen2.5-VL}, which is accessed through its online API.
To construct the Spatial Landmark Knowledge Base (Sec.~\ref{sec:lkb}), we randomly sample 50 trajectories from the training set for each seen scene. For the unseen split, we pre-render images across the environment to serve as scene observations.
The Landmark Recognizer is built upon Qwen-VL-Max \cite{Qwen2.5-VL} and accessed via online APIs.
Unless otherwise specified, we set the lookahead parameter $N_{ahead}$ in the ELG to 2.
For validation, the agent is initialized at the starting point of each episode. At each step, it selects one of six discrete actions: \textit{Move Forward}, \textit{Turn Left}, \textit{Turn Right}, \textit{Ascend}, \textit{Descend}, or \textit{Stop}, along with the number of times the selected action should be executed.
Each action moves or rotates the agent as follows: \textit{Turn Left/Right} 15 degrees, \textit{Ascend/Descend} 2 meters vertically, and \textit{Move Forward} 5 meters.
Navigation ends when the agent chooses the \textit{Stop} action or reaches the maximum number of steps.

\subsection{Experimental Results}
\subsubsection{Quantitative Evaluation}
Table~\ref{tab:overall-performance-l} presents the comparative results on the AerialVLN benchmark. Our \textit{training-free} LookasideVLN surpasses traditional \textit{learning-based} methods such as Seq2Seq and CMA by a clear margin and achieves performance comparable to Zhao’25 on the ``Seen'' split.
Notably, unlike Zhao’25 and other learning-based approaches, our method exhibits strong generalization capability, achieving higher performance on the ``Unseen'' split. This underscores the robustness and generalizability of our zero-shot formulation.

Table~\ref{tab:overall-performance} shows the comparison results on the AerialVLN-S benchmark. LookasideVLN, based on the relatively small Qwen2.5-VL-72B model, outperforms prior methods in terms of success rate (SR) on both ``Seen'' and ``Unseen'' splits.
Several observations can be drawn from the results:
\textup{1)} Statistical methods like Random and Action Sampling perform worst, highlighting the large solution space and action distribution mismatch between training and validation.
\textup{2)} Learning-based methods perform poorly in this challenging setting. Most of them suffer a substantial performance drop in the ``Unseen'' set, suggesting limited generalizability, which is likely due to insufficient training data.
\textup{3)} Zero-shot indoor VLN methods perform poorly in the Aerial VLN task, emphasizing the greater complexity and distinct challenges of outdoor aerial environments compared to indoor scenes.
\textup{4)} Despite using a smaller MLLM, LookasideVLN surpasses previous state-of-the-art Aerial VLN methods on most key metrics. This demonstrates the effectiveness of the Egocentric Lookaside Graph and the path planning strategy based on the Lookaside MLLM Agent.

\begin{figure}[t]
    \centering
    \includegraphics[width=1.0\linewidth]{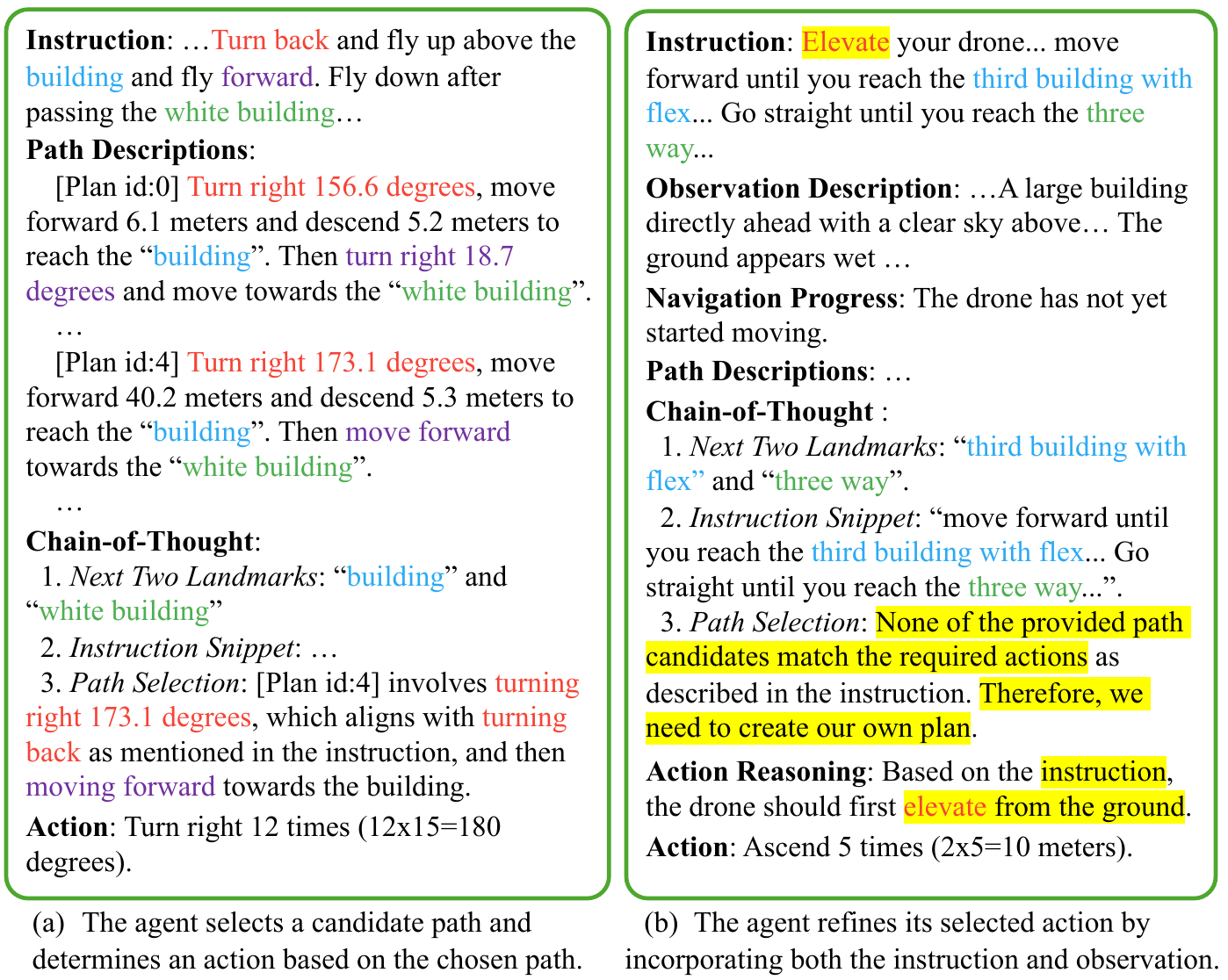}
    \caption{Examples of the agent’s action selection process.}
    \label{fig:cot-example}
\end{figure}
\subsubsection{Qualitative Evaluation}
Fig.~\ref{fig:cot-example} presents two examples of the MLLM Agent planning results.
In Fig.~\ref{fig:cot-example}~(a), the agent selects the candidate path with a relatively large turning angle, aligning with the instruction ``turn back.''
In Fig.~\ref{fig:cot-example}~(b), the agent attempts to find a matching path, but none of the candidates align with the instruction. Recognizing the cue ``elevate'' and observing that it is on the ground, the agent chooses the \textit{Ascend} action instead.
These examples demonstrate that our LookasideVLN can determine actions either through path selection or by aligning with the instruction and current observations.

\subsection{Ablation Study}
To comprehensively analyze the performance of LookasideVLN, we conduct ablation studies on its key components, critical hyperparameters, and the choice of MLLM used in our method. All ablation experiments are performed on the validation ``Seen'' split of the AerialVLN-S benchmark.

\subsubsection{Effect of Egocentric Lookaside Planning}
\begin{table}[t]
    \centering
    \begin{tblr}{
        colspec = {ccccc},
        hline{1,Z} = {1pt},
        hline{2} = {0.5pt},
    }
    ELG        & MLLM Agent    & SR$\uparrow$ & SDTW$\uparrow$ & NE$\downarrow$ \\
    \texttimes  & \texttimes & 2.4          & 1.0            & 405.5            \\
    \checkmark  & \texttimes & 13.8          & 4.6            & 81.6            \\
    \checkmark  & \checkmark & 14.7          & 5.4            & 77.1            \\
    \end{tblr}
    \caption{Ablation study on the Egocentric Lookaside Graph~(ELG) and the Lookaside MLLM Navigation Agent.}
    \label{tab:mod-abl}
\end{table}
Table~\ref{tab:mod-abl} presents an ablation study that evaluates the effectiveness of our proposed modules, including the Egocentric Lookaside Graph~(ELG) and the Lookaside MLLM Navigation Agent.
We observe that removing the ELG and replacing the agent's reasoning with direct action prediction results in the lowest performance, indicating the significance of both modules.
Adding ELG alone improves performance, suggesting that the incorporation of lookaside directional cues significantly benefits long-horizon planning.
When both ELG and the agent are included, LookasideVLN achieves the best overall performance, suggesting that incorporating ELG and the agent's reasoning for path planning better support decision-making in Aerial VLN tasks.

\subsubsection{Impact of Lookahead Horizon $N_{ahead}$}
\begin{table}[t]
    \centering
    \begin{tblr}{
        colspec = {c|ccc},
        hline{1,Z} = {1pt},
        hline{2} = {0.5pt},
        cells = {c},
    }
    $N_{ahead}$        & SR$\uparrow$ & SDTW$\uparrow$ & NE$\downarrow$ \\
    1                  & 9.9          & 3.4            & 84.9            \\
    2                  & 14.7         & 5.4            & 77.1            \\
    3                  & 11.7         & 3.9            & 83.9            \\
    \end{tblr}
    \caption{Ablation study on lookahead horizon $N_{ahead}$.}
    \label{tab:n-ahead-abl}
\end{table}
We further investigate the effect of the lookahead horizon $N_{ahead}$, which determines the number of future subgoals are considered when constructing the ELG. As shown in Table~\ref{tab:n-ahead-abl}, setting $N_{ahead}=2$ yields the best overall performance. When $N_{ahead}=1$, the agent does not incorporate future subgoal information and behaves similarly to a step-by-step planning method with no explicit lookahead, limiting its ability to make effective path planning. In contrast, when $N_{ahead}=3$, the resulting path descriptions become longer and more complex, placing a greater burden on the model’s reasoning capability. This suggests that a moderate lookahead horizon strikes a balance between enhanced spatial modeling and reduced reasoning complexity.
Nevertheless, it is worth noting that our LookasideVLN outperforms the state-of-the-art CityNavAgent~\cite{zhang2025citynavagent} even with a small $N_{ahead}=2$, highlighting the surprising effectiveness of the proposed lookaside directional cues.

\subsubsection{Effect of Different MLLMs}
\begin{table}[t]
    \centering
    \begin{tblr}{
        colspec = {c|ccc},
        hline{1,Z} = {1pt},
        hline{2} = {0.5pt},
        cells = {c},
    }
    MLLM                        & SR$\uparrow$ & SDTW$\uparrow$ & NE$\downarrow$ \\
    LLaVA-7B                    & N/A          & N/A            & N/A            \\
    Qwen2.5-VL-7B               & 9.0          & 1.7            & 306.6          \\
    Qwen2.5-VL-32B              & 14.1         & 4.9            & 94.3           \\
    Qwen2.5-VL-72B (ours)       & 14.7         & 5.4            & 77.1           \\
    \end{tblr}
    \caption{Ablation study on different MLLMs.}
    \label{tab:mllm-abl}
\end{table}
We evaluate the impact of different MLLMs on overall performance, as shown in Table~\ref{tab:mllm-abl}.
LLaVA-7B~\cite{liu2024improved} fails to follow prompts and consistently outputs only observation descriptions, likely due to its limited instruction-following capabilities in long-horizon navigation settings.
Qwen2.5-VL-7B, though lightweight, exhibits limited ability to handle the complex long-horizon reasoning required for Aerial VLN. In contrast, both the 32B and 72B versions of Qwen2.5-VL achieve significantly stronger performance. These results suggest that more capable MLLMs provide better support for effective path reasoning.

\section{Conclusion}

In this work, we introduced LookasideVLN, a new Aerial VLN paradigm that leverages directional cues in natural language to enhance spatial reasoning and efficiency. Unlike prior lookahead-based approaches that rely heavily on costly landmark sequence alignment, LookasideVLN integrates an Egocentric Lookaside Graph, a Spatial Landmark Knowledge Base, and a Lookaside MLLM Navigation Agent to achieve lightweight yet context-aware path planning.
Through extensive experiments, we demonstrated that LookasideVLN substantially outperforms the state-of-the-art CityNavAgent, despite using only single-level lookahead planning. These results highlight the value of incorporating directional semantics as a complementary spatial cue in Vision-and-Language Navigation.
\subsection*{Acknowledgments}
This work is supported in part by the National Key
R\&D Program of China (NO.~2024YFB3908503 and 2024YFB3908500), and in part by the
National Natural Science Foundation of China (NO.~62322608).

{
    \small
    \bibliographystyle{ieeenat_fullname}
    \bibliography{main}
}

\clearpage
\setcounter{page}{1}
\setcounter{section}{0}
\maketitlesupplementary

\section{Spatial Landmark Knowledge Base}
\subsection{Landmark Recognizer}
The Landmark Recognizer aims to identify visible landmarks present in historical observations. To this end, we leverage the strong image captioning capabilities of multimodal large language models to generate landmark descriptions from RGB observations. Specifically, we use Qwen-VL-Max \cite{Qwen2.5-VL} to extract landmark descriptions from each panoramic observation $O = \{o_i\}_{i = 1}^5$ (without the upward view). The prompt used for the Landmark Recognizer is as follows:
\begin{promptbox}[Prompt for Landmark Recognizer]
{\large\textbf{Task}}\\
Analyze the drone's multi-view images and list distinctive static landmarks using concise natural language descriptions. Follow these guidelines:
\begin{enumerate}
\item[1.] \textbf{Key Focus}:
\begin{itemize}
    \item Describe landmarks in several word phrases using this pattern (not strictly): \\
    \texttt{[Color] [Material] [Type] [Distinctive Feature]} \\
    (e.g., \textit{``red brick water tower''}, \textit{``blue ladder''})
\end{itemize}
\item[2.] \textbf{View Processing}:\\
For each view (`front/bottom/left/right/back`):
\begin{itemize}
    \item List navigation-critical objects, empty if none.
    \item Use comparative references when helpful: \\
    (e.g., \textit{``taller than surrounding buildings''})
\end{itemize}
\end{enumerate}
{\large\textbf{Example Output (JSON)}}
\begin{tcolorbox}[
    colframe=gray!15,
    opacityback=0,
    left=0pt,
    right=0pt,
    top=0pt,
    bottom=0pt
]
\begin{tabular}{ll}
    ``front'': [\textit{``gray metal tower''}, \textit{``blue bridge''}],\\
    ``bottom": [\textit{``red-roofed circular building''}],\\
    ``left'': [\textit{``green glass skyscraper''}],\\
    ``right'': [\textit{``gray dome-shaped observatory''}],\\
    ``back'': []
\end{tabular}
\end{tcolorbox}
\rule{\linewidth}{0.4pt}
{\large\textbf{Input}}\\
The drone's multi-view images are:
\begin{itemize}
    \item Front View: $o_1$,
    \item Bottom View: $o_2$,
    \item Left View: $o_3$,
    \item Right View: $o_4$,
    \item Back View: $o_5$
\end{itemize}
\end{promptbox}

\subsection{Extracting Landmark Descriptions}
To obtain the landmark descriptions $\{l^{instr}_i\}_{i=1}^{N}$ used in Sec.~\ref{sec:lkb-retrieval} of the main paper, we develop a Landmark Parser that directly extracts landmark descriptions from the instruction~$\mathcal{I}$ while preserving their original order.
We implement the Landmark Parser using the large language model Qwen-Max \cite{qwen2025qwen25technicalreport}. The prompt used for the Landmark Parser is as follows:
\begin{promptbox}[Prompt for Landmark Parser]
{\large\textbf{Task}}\\
You are a \textbf{Navigation Landmark Parser}. Your job is to:
\begin{enumerate}
\item[1.] \textbf{Identify every landmark reference} in the instruction, including:
\begin{itemize}
   \item \textbf{Street names}\\
   (e.g., \textit{``Main Street''}, \textit{``5th Avenue''})
   \item \textbf{Road types}\\
   (e.g., \textit{``highway''}, \textit{``dirt road''})  
   \item \textbf{Intersections}\\
   (e.g., \textit{``the corner of Park and Elm''})  
   \item \textbf{Points of interest}\\
   (e.g., \textit{``gas station''}, \textit{``red mailbox''})
\end{itemize}
\item[2.] \textbf{List them EXACTLY in order}\\
including all prefixes/suffixes.
\end{enumerate}
{\large\textbf{Output Format (JSON List)}}
\begin{tcolorbox}[
    colframe=gray!15,
    opacityback=0,
    left=0pt,
    right=0pt,
    top=0pt,
    bottom=0pt
]
[``landmark\_name\_1'', ``landmark\_name\_2'', ...]
\end{tcolorbox}
{\large\textbf{Examples}}
\begin{itemize}
    \item Instruction: \textit{``Turn left on Maple Street, then right at the bank.''}\\
    Output: [\textit{``Maple Street''}, \textit{``bank''}]
    \item Instruction: \textit{``Follow Highway 1 until the second traffic light.''}\\
    Output: [\textit{``Highway 1''}, \textit{``traffic light''}]
\end{itemize}
\rule{\linewidth}{0.4pt}
{\large\textbf{Input}}\\
The navigation instruction to process is `` $\mathcal{I}$ ''.
\end{promptbox}

\subsection{Landmark Localization with Bounding Box}
In Eq.~\ref{eq:landmark_localization}, we use the center pixel coordinates $(u_i, v_i)$ of the bounding box $b_i$, along with the estimated depth $\bar{d}_i$, to compute the 3D position of the landmark $l_i$.
We apply perspective projection to our depth observations to ensure consistency with the RGB inputs. In the following paragraphs, we detail the transformation from pixel coordinates to world coordinates under perspective projection, as it involves more complex computations than those under orthographic projection.

\begin{figure}
    \centering
    \includegraphics[width=1.0\linewidth]{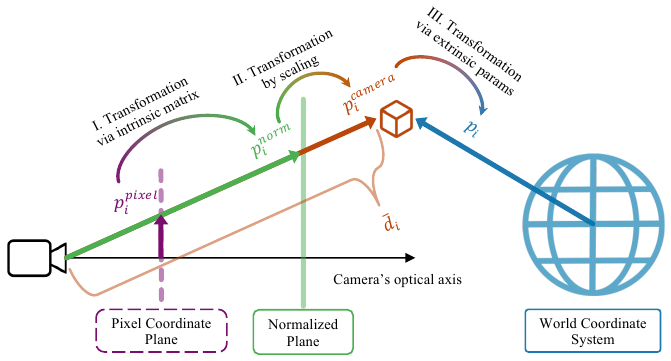}
    \caption{Illustration of the transformation from pixel coordinates to world coordinates under perspective projection}
    \label{fig:coord_trans}
\end{figure}

As shown in Fig.~\ref{fig:coord_trans}, given the homogeneous pixel coordinate $p^{pixel}_i = [u_i, v_i, 1]^\top$, the corresponding position in normalized plane\footnote{The normalized image plane is a virtual plane placed at a unit distance in front of the camera and perpendicular to its optical axis.} is computed as:
\begin{equation}
    p^{norm}_i = K^{-1} p^{pixel}_i,
\end{equation}
where $K \in \mathbb{R}^{3 \times 3}$ denotes the camera's intrinsic matrix. The depth value $\bar{d}_i$ of the target landmark is defined as the Euclidean distance from the camera center to the object under perspective projection. Accordingly, the landmark’s 3D position in camera coordinates can be represented as:
\begin{equation}
    p^{camera}_i = \bar{d}_i \cdot \frac{p^{norm}_i}{||p^{norm}_i||_2}.
\end{equation}
To transform this point into world coordinates, the camera's extrinsic parameters are applied:
\begin{equation} \begin{aligned}
    p_i &= R p^{camera}_i + T \\
    &= \frac{\bar{d}_i}{\|K^{-1} p^{pixel}_i\|_2} \cdot R K^{-1} p^{pixel}_i + T,
\end{aligned} \end{equation}
where $R \in SO(3)$ and $T \in \mathbb{R}^3$ are the camera's rotation matrix and translation vector, respectively.

\subsection{Pruning Strategy for SLKB}
Since the agent may revisit the same location multiple times throughout its navigation history, the recorded observations can be overlapped. Thus, multiple detection results of the same landmark in nearby locations may appear in the Spatial Landmark Knowledge Base (SLKB)---\ie, several locations in $\{p^{kb}_{i,0}, p^{kb}_{i,1}, \dots\}$ corresponding to the landmark description $l^{kb}_i$ may refer to the same physical instance in the scene. To reduce redundancy, we apply a non-maximum suppression (NMS)-like pruning strategy based on the Euclidean distances between candidate locations. Specifically, for the candidate location list $\{p^{kb}_{i,0}, p^{kb}_{i,1}, \dots\}$ of each landmark description $l^{kb}_i$ in the SLKB, we first sort the candidates by their associated confidence scores in descending order. Starting from the highest-scoring location, we select it as a representative detection and remove all other candidates within a predefined Euclidean distance threshold of 20 units, as these are considered duplicates. This process is repeated iteratively on the remaining candidates until no overlapping locations remain. The result is a refined set of landmark positions with minimal spatial redundancy, improving the accuracy and efficiency of the knowledge base.

\section{Pruning Strategy for ELG}
Since large-scale urban scenes often contain many visually similar landmark instances that can be grounded by the same landmark description, traversing the entire graph can lead to redundancy. To address this, we adopt a pruning strategy to eliminate redundant nodes and edges in the ELG. Specifically, given the ordered set of unvisited landmarks and their candidate positions, $\mathcal{L}^{unvis} = \big\{ l^{unvis}_i: \{ p^{unvis}_{i,0}, p^{unvis}_{i,1}, \dots \} \big\}_{i=1}^{N_{ahead}}$, we retain the top $N_{next} = 6$ candidate positions closest to the agent's current location for the first unvisited landmark $l^{unvis}_1$.
For each candidate position $p^{unvis}_{i,j}$ of the $i$-th unvisited landmark $l^{unvis}_i$, we connect it to the $N_{subseq} = 2$ nearest candidate positions of the subsequent landmark $l^{unvis}_{i+1}$, i.e., nearest positions selected from $\{ p^{unvis}_{i+1,m} \}_{m=1}^{\dots}$.

\section{Implementation of the Lookaside MLLM Navigation Agent}
We build the agent with a chain-of-thought mechanism for robust and explainable path planning.
Specifically, we design three types of prompts to handle different situations the agent may encounter during navigation:
    \textup{1)} \textit{When the Egocentric Lookaside Graph (ELG) is available}, the agent selects a navigation path by reasoning over future subgoals and their spatial relationships.
    \textup{2)} \textit{When a path has been selected and the next candidate landmark is known}, the agent determines the next action based on the landmark's location. This action is further refined using the current observation and the navigation instruction.
    \textup{3)} \textit{In rare cases where the instruction contains no identifiable landmark}, the agent navigates by relying solely on visual observations and the navigation instruction.
For each of these situations, we craft dedicated prompts for the agent to enable effective and robust reasoning. Moreover, the agent maintains a record of visited landmarks and summarizes its navigation history in text to support contextual reasoning. Below, we outline the prompt strategies for each scenario.

When the ELG is available, the agent is guided to reason over candidate paths and predict the next action accordingly. The prompt used at this stage is as follows:
\begin{promptbox}[Prompt with ELG]
{\large\textbf{System Message}}\\
You are a visual-and-language navigation agent that follows navigation instructions to move to each specific landmark and finally reach the destination.

{\large\textbf{Input Description}}
\begin{itemize}
    \item \textbf{Instruction}: A natural language navigation instruction that guides the agent to reach some specific landmarks.
    \item \textbf{History}: A record of the landmarks that the agent has visited with a brief trajectory description.
    \item \textbf{Visited Landmark}: A list of landmarks that the agent has visited.
    \item \textbf{Candidate Paths}: A set of candidate navigation paths.
    \item \textbf{Current Views}: Six views observed at the current time step: front, bird’s-eye, left, right, back, and upward views.
\end{itemize}
\rule{\linewidth}{0.4pt}
{\large\textbf{Output Description}}
\begin{itemize}
    \item \textbf{Observation Description}: The current observation and the landmark you seem based on the current views.
    \item \textbf{Navigation Progress}: Describe the current navigation progress based on the instruction and the history.
    \item \textbf{Reasoning for Path Selection}: Identify unvisited landmarks in the Candidate Paths to visit next. Give a description on how to reach the unvisited landmarks based on the instruction step by step with the relevant instruction snippets. For example, ``Next Two Landmarks: A and B. Instruction Snippet: \textit{Turn left at A, then move forward to B}.''
    \item \textbf{Selected Path ID}: Based on the reasoning above, select one of the candidate paths to follow by providing its ID. If you choose to create your own plan, respond with -1.
    \item \textbf{Action Reasoning}: Given the selected path, determine the next action toward the next landmark. Refine your decision using the navigation instruction and current visual observation.
    \item \textbf{Action}: Choose the next action from \{\textit{forward}, \textit{turn\_left}, \textit{turn\_right}, \textit{ascend}, \textit{descend}, \textit{stop}\} The \textit{forward} action will let you move \textbf{5 meters} in current direction; A turning action will result in a \textbf{15 degree} turning; The \textit{ascend} / \textit{descend} action will cause an altitude change of \textbf{2 meters}; The stop action should be triggered when the final landmark is reached.
    \item \textbf{Calculation for Number of Executions}: For example, executing a turning action six times at 15 degrees per turn will result in a total rotation of 15 x 6 = 90 degrees.
    \item \textbf{Number of Executions}: The number of times this action should be executed.
    \item \textbf{Updated History}: Update the history according to the landmark you see and the action you take.
\end{itemize}
\end{promptbox}

After selecting a path, the agent determines the next action based on the spatial relationship to the upcoming landmark, further refined using current observations and instructions. The prompt at this stage is adapted from the previous version by replacing ``Candidate Paths'' with ``The Next Landmark'' in the input. Correspondingly, the path selection reasoning is replaced with landmark-based reasoning. The prompt used at this stage is as follows:
\begin{promptbox}[Prompt with the next landmark]
{\large\textbf{System Message}} ...\\
{\large\textbf{Input Description}}
\begin{itemize}
    \item[] ...
    \item \textbf{The Next Landmark}: The upcoming landmark the agent needs to reach, along with its relative location.
    \item[] ...
\end{itemize}
\rule{\linewidth}{0.4pt}
{\large\textbf{Output Description}}
\begin{itemize}
    \item[] ...
    \item \textbf{Reasoning for Landmark to Follow}: Reason whether to proceed toward the next landmark by considering the given instruction, current observation, and the landmark's relative position.
    \item \textbf{Whether to Follow the Next Landmark}: Select whether to follow the next landmark or not. Respond with \textit{follow} or \textit{not follow}.
    \item[] ...
\end{itemize}
\end{promptbox}

In rare cases where instructions lack identifiable landmarks, the agent relies on general spatial understanding to determine navigation direction. Unlike previous prompts, information about candidate paths and landmark locations is excluded from the input.

\section{More Experiment Results}

\subsection{Additional MLLM ablations.}
\begin{table}[h]
    \centering
    \caption{Additional ablation study on different MLLMs.}
    \small
    \begin{tblr}{
        colspec = {cc|ccc},
        hline{1,Z} = {1pt},
        hline{2} = {0.5pt},
        row{1} = {font=\bfseries},
    }
    Category    & Method & SR$\uparrow$ & SDTW$\uparrow$ & NE$\downarrow$ \\
    \SetCell[r=4]{m} Qwens
        & Qwen-2.5-VL-7B & 9.0 & 1.7 & 306.6 \\
        & Qwen-3-VL-8B & 11.7 & 3.8 & 114.1 \\
        \hline
        & Qwen-2.5-VL-32B & 14.1 & 4.9 & 94.3 \\
        & Qwen-3-VL-32B & 11.7 & 5.1 & 84.8 \\
    \hline
    \SetCell[r=2]{m} GPTs
        & GPT-4V & 15.0 & 6.6 & 80.8 \\
        & GPT-4o & 15.3  & 6.3 & 83.5 \\
    \end{tblr}
    \label{tab:more-mllms}
\end{table}
We add GPT-series and Qwen3VL as MLLM backbones in Tab.~\ref{tab:more-mllms}. We note that Qwen2.5VL-32B is specifically optimized with RL\footnote{Refer to https://qwen.ai/blog?id=qwen2.5-vl-32b}, which explains its stronger performance compared to Qwen3VL-32B, while GPTs achieve the best overall results, further validating the effectiveness of LookasideVLN.

\subsection{Sensitivity analysis on depth noise.}
We conduct a sensitivity analysis in Tab.~\ref{tab:depth-noise-ana} by adding Gaussian noise to the depth $\bar{d}_i$ in Eq.~3. LookasideVLN remains robust under large noise (up to $\sigma=5$ meters), as the agent can still infer landmark directions from its observations.
\begin{table}[h]
    \centering
    \caption{Sensitivity analysis on the seen split of AerialVLN-S.}
    \small
    \begin{tblr}{
        colspec = {c|ccc},
        hline{1,Z} = {1pt},
        hline{2} = {0.5pt},
        row{1} = {font=\bfseries},
    }
    Depth Noise Std. ($\sigma$) & SR$\uparrow$ & SDTW$\uparrow$ & NE$\downarrow$ \\
        $\sigma = 0$ meters     & 14.7 & 5.4 & 77.1 \\
        $\sigma = 5$ meters     & 13.2 & 4.7 & 83.5 \\
        $\sigma = 20$ meters    & 11.1 & 3.5 & 84.4 \\
    \end{tblr}
    \label{tab:depth-noise-ana}
\end{table}

\begin{figure*}[!t]
    \centering
    \includegraphics[width=1.0\linewidth]{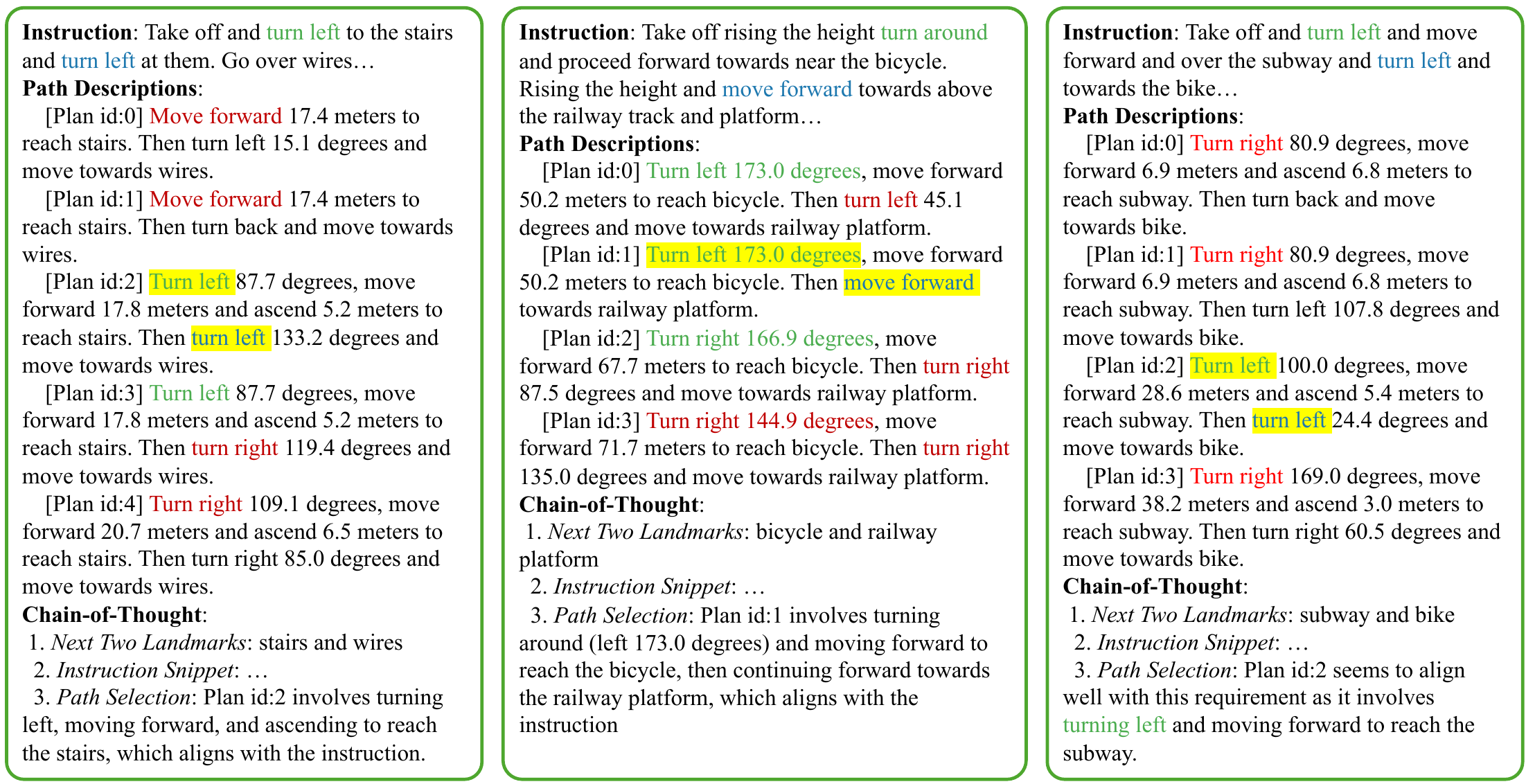}
    \caption{Examples of the agent's reasoning. The agent selects optimal paths using descriptions derived from the Egocentric Lookaside Graph. Highlighted words indicate successful path decisions that align with directional cues in the instructions.}
    \label{fig:more-cot}
\end{figure*}

\begin{figure*}[t]
    \centering
    \includegraphics[width=1.0\linewidth]{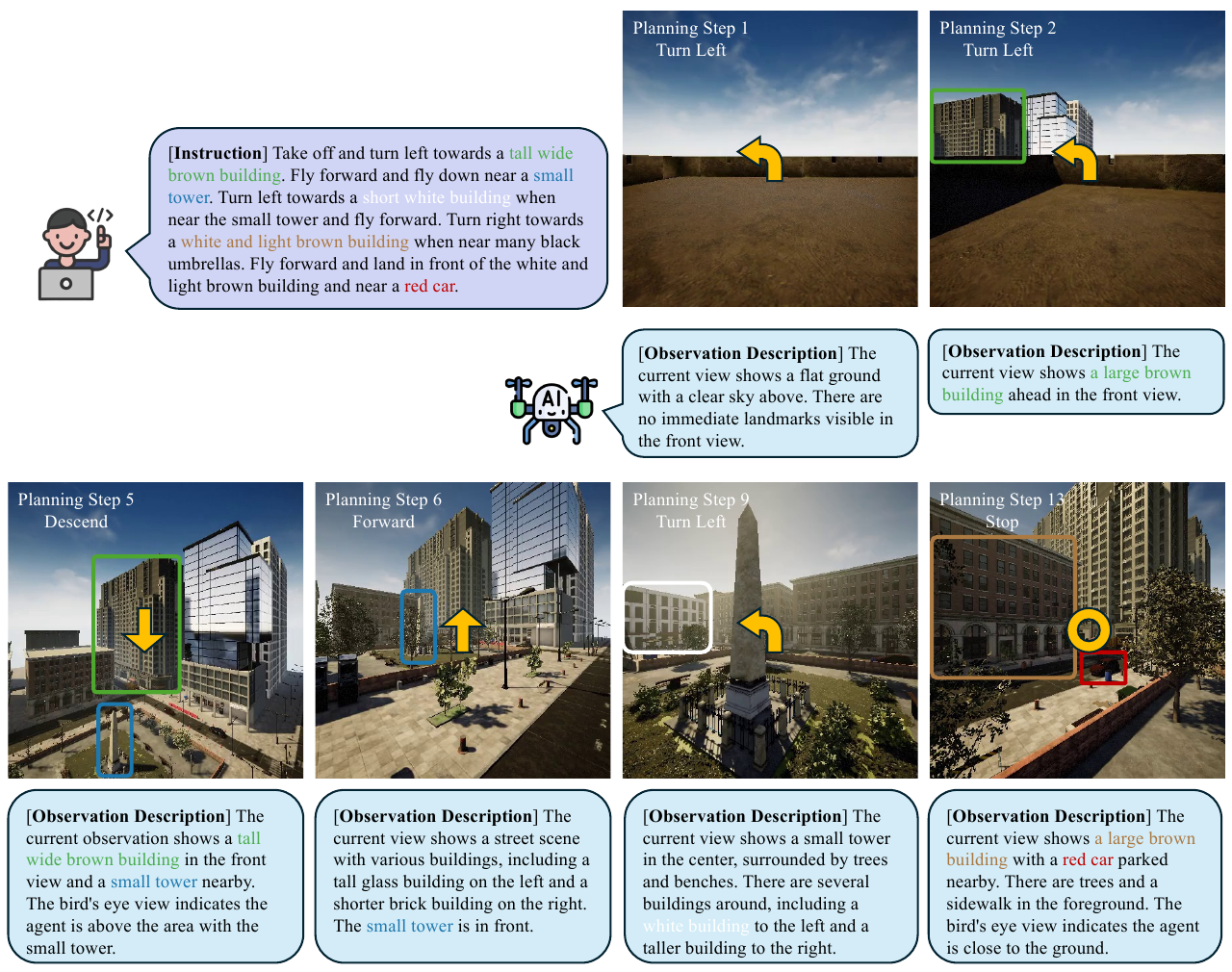}
    \caption{Visualization of key steps in a navigation episode. LookasideVLN provides accurate observation descriptions and makes appropriate action decisions.}
    \label{fig:nav-vis}
\end{figure*}

\subsection{Qualitative Examples of Reasoning Process}
Fig.~\ref{fig:more-cot} shows more examples of the chain-of-thought reasoning. The agent successfully grounds instructions with path descriptions derived from the ELG and leverages the egocentric lookaside directional relationships for robust path selections.

Fig.~\ref{fig:nav-vis} shows key steps of a navigation episode where LookasideVLN follows complex instructions through an urban environment. Starting with textual instructions, the agent successfully navigates through multiple waypoints, maintaining correct visual perceptions. The sequence demonstrates the agent's ability to interpret visual scenes and make appropriate navigation decisions to reach the target destination.

\subsection{Failure Cases}
\begin{figure}[!t]
    \centering
    \includegraphics[width=1.0\linewidth]{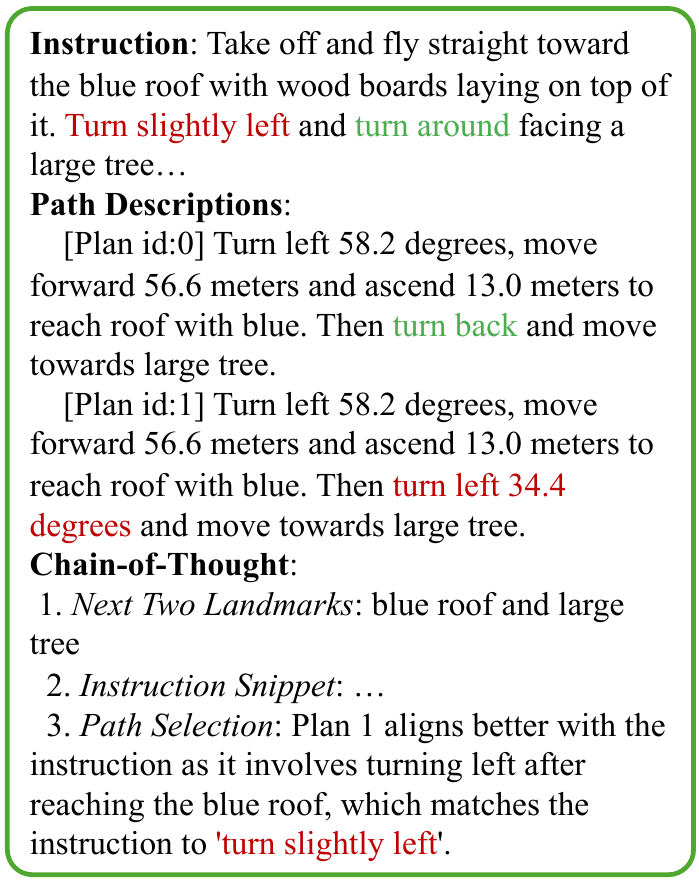}
    \caption{An example of failure cases where the agent's reasoning process fails due to hallucinations. The agent correctly identifies the directional cue ``slightly left'' but fails to recognize the subsequent ``turn around'' instruction, leading to failed path planning.}
    \label{fig:failed-cot}
\end{figure}
Fig.~\ref{fig:failed-cot} presents a failure case in the reasoning process. The instruction includes two consecutive directional cues---``\textit{turn slightly left}'' and ``\textit{turn around}''. The agent is misled by the former and fails to attend to the latter, resulting in an incorrect path selection.

\section{Limitations and Future Work}
Although LookasideVLN achieves state-of-the-art performance in the simulated environment, it has not yet been deployed in the real world. Future work may focus on bridging the sim-to-real gap and evaluating the system on real UAVs. Additionally, improving robustness under real-world conditions will be an important direction. A detailed discussion of the limitations and future work is provided below.

\subsection{Limitation Discussion}
Although LookasideVLN outperforms most prior state-of-the-art baselines on the challenging AerialVLN benchmark, the overall performance remains limited. The reasons are twofold:
\begin{enumerate}
    \item Learning-based baselines are constrained by the scarcity of annotated trajectories and the limited scale of training scenes. While they can perform accurate action selection (at each step) during end-to-end training, they tend to accumulate errors (over the entire navigation process) in evaluation and struggle to generalize to unseen environments. This issue may be alleviated through more cost-efficient data collection strategies and dedicated error-correction mechanisms.
    \item Zero-shot LLM-based approaches are strong in understanding natural language instructions and visual observations. However, they struggle to comprehend structured 3D scenes when relying solely on language and RGB inputs, lacking access to fine-grained 3D information such as depth cues. They also face challenges in ego-motion understanding---often misinterpreting ego-trajectories when processing egocentric navigation videos. These limitations may be mitigated by leveraging more advanced MLLMs specialized in 3D scene comprehension and ego-motion modeling.
\end{enumerate}

\subsection{Future Work}
We validate LookasideVLN in simulated environments but have not yet conducted real-world evaluations. In the AerialVLN benchmark, the agent is instructed to output a single action at each step to align with the benchmark’s evaluation settings. However, invoking the MLLM-based agent for reasoning at every step is computationally expensive and cannot be directly deployed on real-world UAVs. In practical deployment, prompting the agent to generate planned waypoints along the selected landmark-level path within the ELG---and executing these waypoints using low-level motion planners---could reduce reasoning time while enabling safety checks and occlusion avoidance.

\end{document}